%% file: root.tex
\documentclass[10pt,twocolumn,letterpaper]{article}

\usepackage{wacv}

\wacvfinalcopy

\usepackage{svg}
\usepackage{amsmath} 
\usepackage{amssymb}  
\usepackage{tabularx}
\usepackage[utf8]{inputenc}
\usepackage[T1]{fontenc}
\usepackage{textcomp}
\usepackage{gensymb}
\usepackage{multirow}
\usepackage{paralist}
 
\usepackage{booktabs}
\usepackage{xcolor}
\usepackage{graphicx}
\graphicspath{ {./images/}}
\usepackage{float}
\usepackage{capt-of}
\usepackage{adjustbox}
\usepackage[colorlinks=true, pagebackref]{hyperref}
\usepackage[font=small]{caption}


\usepackage{cite}

\newcommand{\monolayout}{\emph{MonoLayout}}
\newcommand{\bev}{bird's eye view}
\newcommand{\mb}[1]{\mathbf{#1}}

\title{\textbf{\monolayout{}: Amodal scene layout from a single image}}

\usepackage{authblk}

\makeatletter
\renewcommand\AB@affilsepx{, \protect\Affilfont}
\makeatother

\author[1]{Kaustubh Mani\thanks{Corresponding author: \url{kaustubh3095@gmail.com} \\ Project page: \url{https://hbutsuak95.github.io/monolayout}}}
\author[1]{Swapnil Daga}
\author[2]{Shubhika Garg}
\author[1]{N. Sai Shankar}
\author[3,4]{Krishna Murthy Jatavallabhula \thanks{Both (un)supervisors procrastinated equally}}
\author[1]{K. Madhava Krishna$^\dagger$}

\affil[1]{Robotics Research Center, IIIT Hyderabad}
\affil[2]{IIT Kharagpur}
\affil[3]{Mila - Quebec AI Institute, Montreal}
\affil[4]{Universit\'e de Montr\'eal}

\begin{document}

\makeatletter
\let\@oldmaketitle\@maketitle
\renewcommand{\@maketitle}{\@oldmaketitle
\centering
\vspace{-4cm}
\includegraphics[scale=0.5]{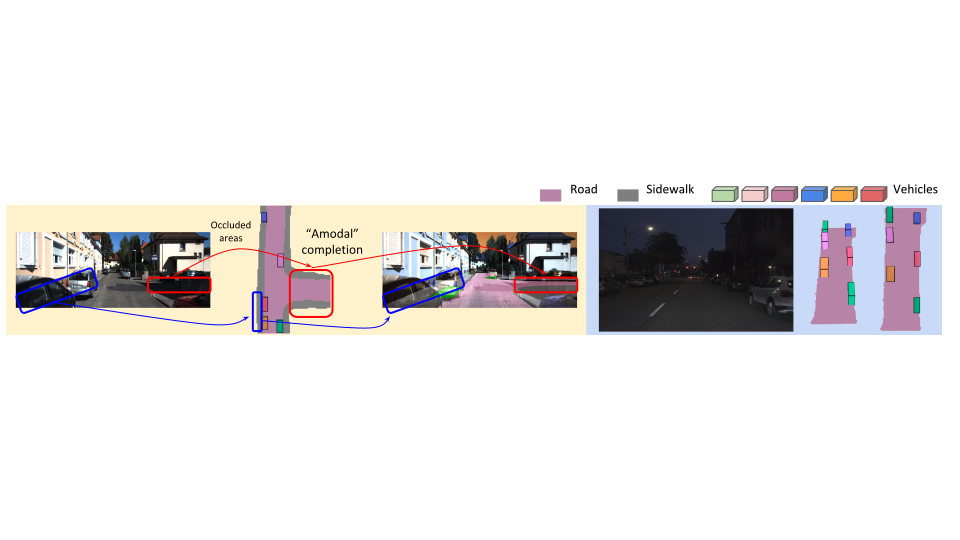}
\vspace{-3.8cm}
\captionof{figure}{
\textbf{\monolayout{}}: Given only a single image of a road scene, we propose a neural network architecture that reasons about the \emph{amodal} scene layout in \bev{} in \emph{real-time} (30 fps). Our approach, dubbed \monolayout{} can \emph{hallucinate} regions of the static scene (road, sidewalks)---and traffic participants---that do not even project to the visible regime of the image plane. Shown above are example images from the KITTI \cite{KITTI} (left) and Argoverse \cite{chang2019argoverse} (right) datasets. \monolayout{} outperforms prior art (by more than a 20\% margin) on hallucinating occluded regions.
}
\label{fig:teaser_new}
\vspace{0.2cm}
}
\makeatother

\maketitle


\begin{abstract}
\vspace{-0.3cm}


In this paper, we address the novel, highly challenging problem of estimating the layout of a complex urban driving scenario. Given a single color image captured from a driving platform, we aim to predict the \bev{} layout of the road and other traffic participants. The estimated layout should reason beyond what is visible in the image, and compensate for the loss of 3D information due to projection. We dub this problem \emph{amodal scene layout estimation}, which involves \emph{hallucinating} scene layout for even parts of the world that are occluded in the image. To this end, we present \monolayout{}, a deep neural network for real-time amodal scene layout estimation from a single image.
We represent scene layout as a multi-channel semantic occupancy grid, and leverage adversarial feature learning to ``hallucinate" plausible completions for occluded image parts.
We extend several state-of-the-art approaches for road-layout estimation and vehicle occupancy estimation in \bev{} to the amodal setup for rigorous evaluation. By leveraging temporal sensor fusion to generate training labels, we significantly outperform current art over a number of datasets.
A video abstract of this paper is available \href{https://www.youtube.com/watch?v=HcroGyo6yRQ}{here}.

\end{abstract}


\input{text/intro}

\input{text/relatedwork}

\input{text/approach}

\input{text/results}

\input{text/conclusions}
\input{text/acknowledgements}

{\small
\bibliographystyle{ieee}
\bibliography{references}
}


\appendix

\input{text/appendices}



\end{document}


\maketitle

\section{Title}

\subsection{Timing analysis}
 
We also show the computation test time of our method as compared to similar methods in Table \ref{table:computation_time_table}. Our network does not require discriminator to be used during inference time. It achieves real time inference rate of approx. $30$ Hz for an input image with a resolution of 512 * 512 pixels and an output map with 128 * 128 occupancy grid cells using a Nvidia GeForce GTX 1080Ti GPU. The code for \cite{schulter2018learning} is not publicly available, and the computation time is based on the PSMNet \cite{psmnet} backbone they use. Here again the proposed method is almost an order faster than previous methods making it more attractive for on-road implementations.

\setlength\extrarowheight{5pt}
\begin{table}[!hbt]
	\centering
	
    \begin{adjustbox}{max width=\linewidth}
    \begin{tabular}{|c|c|c|}
		\hline
        \textbf{Method} & \textbf{Parameters} & \textbf{Computation Time} \\
        \hline
        OFT\cite{roddick2018orthographic} & $24.5$ M & $2$ fps \\
        \hline
        MonoOccupancy\cite{lu2019monocular} & $10.4$ M & $34$ fps \\
        \hline
        Schulter et al.\cite{schulter2018learning} & $>> 20$ M & $< 3$ fps \\
        \hline
        \monolayout{} (Ours) & 19.6M & 30 fps \\
        \hline
	\end{tabular}
    \end{adjustbox}
    \caption{A comparative study of test computation time on NVIDIA GeForce GTX 1080Ti GPU for different methods on the images of KITTI\cite{KITTI} RAW dataset.}
        
    \label{table:computation_time_table}
\end{table}
\begin{figure}[!t]
    \centering
    \includegraphics[height=5cm, width=0.5\textwidth]{legend3rgb.png}
    \caption{Caption}
    \label{fig:my_label}
\end{figure}

\subsection{Trajectory Generation}

One of the use-cases of \monolayout{} can be to generate trajectories from the obtained \emph{amodal} scene layout. Methods such as INFER\cite{srikanth2019infer} predict future trajectories from occupancy grid information about road, vehicle, target, obstacles and lane. Input is generated in INFER\cite{srikanth2019infer} from instance segmentation, semantic segmentation and depth information. We replace their target, road and vehicle occupancy grid inputs with our road and vehicle layout in \bev{} that is generated from \monolayout{}. We not only generate at par results with INFER\cite{srikanth2019infer} but also reduce the computation time by replacing mutiple networks in INFER\cite{srikanth2019infer} with a single network for intermediate representation. The model was trained on 128 * 128 occupancy grid representing 40m * 40m area in real world. Further, we propose an extension to generate accurate tracklets by leveraging our vehicle occupancy detection results. Upon the generated vehicle occupancies, traditional instance segmentation methods like blob detection are applied to obtain the coordinates of the center of vehicle in the bird's eye view. We then associate the vehicles through frames, by calculating IoU values for different vehicle blobs between frames, thus enabling us to generate tracklets efficiently. Table \ref{table:tracking_table} compares the method described above against the method described in INFER\cite{srikanth2019infer} in which external information like instance segmentation and depth maps are used in order to get their 3D coordinates and correspond them in sebsequent frames. Table \ref{table:tracking_table} clearly shows we outperform the baseline method used for tracking vehicles and thus proving that better tracklets can be generated in bird's eye view using our car occuapncy layout without requiring any extra information.   

\begin{figure}[!t]
    \centering
    \includegraphics[height=5cm,width=0.5\textwidth]{legend3.png}
    \caption{Caption}
    \label{fig:my_label}
\end{figure}

\setlength\extrarowheight{5pt}
\begin{table}[!hbt]
	\centering
    
    \begin{adjustbox}{max width=\linewidth}
		\begin{tabular}{|c|c|c|c|}
		\hline
		\textbf{Method} & \textbf{Mean Error in Z (m)} & \textbf{Mean Error in X (m)} & \textbf{Mean L2 error (m)}\\
        \hline
        Infer & $1.08$ & $0.51$ & $1.27$ \\
        \hline
        Monolayout & $\mb{0.23}$ & $\mb{0.47}$ & $\mb{0.58}$ \\
        \hline

	    \end{tabular}
    \end{adjustbox}
    \caption{Quantative results: We show the preformance of vehicle tracking first using instance segmentation and depth maps on RBG images as descibed in INFER and second using traditional instance segmenation techniques like blob detection on MonoLayout's Car occupancy which is then compared against the actual ground truth tracklets available in Kitti Tracking dataset}
    \label{table:tracking_table}
\end{table}

\subsection{Metrics}

\subsubsection*{Road layout estimation}
To evaluate estimated road layouts, we use intersection-over-union (IoU) as our primary metric. We split IoU evaluation into two parts and measure IoU for the entire static scene, as well as IoU for occluded regions (i.e., regions of the road that are occluded in the image and were hallucinated by \monolayout{}).

\subsubsection*{Vehicle occupancy estimation}
While most approaches to vehicle detection evaluate only mean Average Precision (mAP), it has been shown to be a grossly inaccurate measure of how tight a bounding box is \cite{prod}. We hence adopt mean Intersection-over-Union (mIoU) as our primary basis of evaluation. To ensure a fair comparision with prior art, we also report mAP.\footnote{We outperform existing methods under both these metrics}.

{\small
\bibliographystyle{ieee}
\bibliography{references}
}

%% file: text/intro.tex
\section{Introduction}
\label{sec:introduction}

The advent of autonomous driving platforms has led to several interesting, new avenues in perception and scene understanding. While most of the industrially-led solutions leverage powerful sensors (eg. lidar, precision GPS, etc.), an interesting research question is to push the capabilities of monocular vision sensors. To this end, we consider the novel and highly challenging task of estimating \emph{scene layout} in \bev{}, given only a single color image.

Humans have a remarkable cognitive capability of perceiving \emph{amodal} attributes of objects in an image. For example, upon looking at an image of a vehicle, humans can nominally \emph{infer} the occluded parts, and also the potential geometry of the surroundings of the vehicle.
While modern neural networks outperform humans in image recognition and object detection \cite{he2015delving, cirecsan2012multi, yu2017sketch, rota2018place, he2017mask, girshick2015fast, ren2015faster}, they still lack this innate cognitive capability of reasoning beyond image evidence. With this motivation, we propose \monolayout{}, a neural architecture that takes as input a color image of a road scene, and outputs the \emph{amodal} scene layout in \bev{}. \monolayout{} maps road regions, sidewalks, as well as regions occupied by other traffic participants such as cars, to \bev{}, in a single pass, leveraging adversarial feature learning.

\begin{figure*}[!ht]
  \centering
  \vspace{-3.5cm}
  \includegraphics[width=\textwidth]{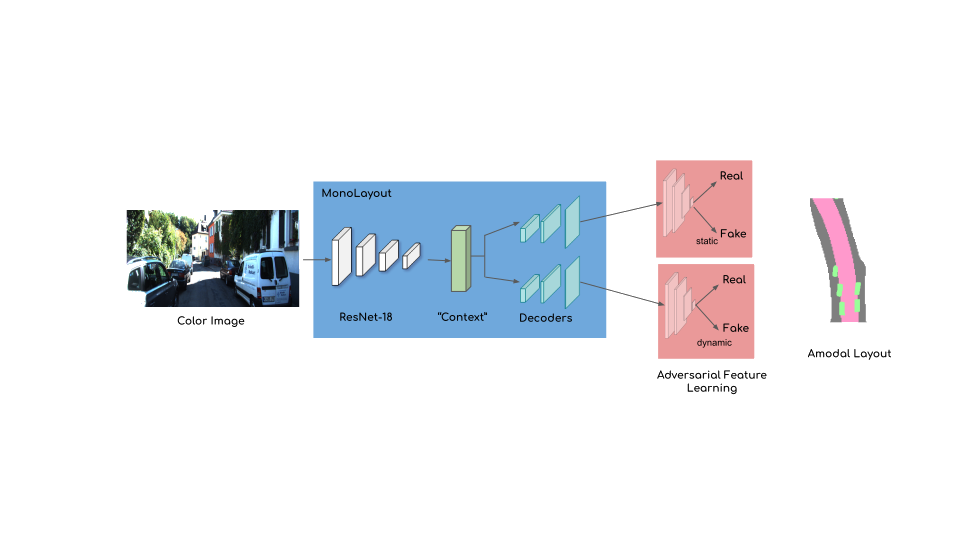}
  \vspace{-3.5cm}
  \caption{\textbf{Architecture}: \monolayout{} takes in a color image of an urban driving scenario, and predicts an amodal scene layout in \bev{}. The architecture comprises a \emph{context encoder}, \emph{amodal layout decoders}, and \emph{two discriminators}.}
  \label{fig: main_architecture}
\end{figure*}

To the best of our knowledge, \monolayout{} is the first approach to \emph{amodally} reason about the static and dynamic objects in a scene.
We show that, by using a shared context to reason about scene entities, \emph{MonoLayout} achieves better performance on each task when compared to approaches that train only for a particular task.
On the task of amodal scene layout estimation, \monolayout{} outperforms all evaluated baselines by a significant margin on several subsets of the KITTI \cite{KITTI} and the Argoverse \cite{chang2019argoverse} datasets. Further, \monolayout{} achieves state-of-the-art object detection performance in \bev{}, without using any form of thresholding / postprocessing. In summary, our contributions are the following:

\begin{compactenum}
    \item We propose \monolayout{}, a practically motivated deep architecture to estimate the \emph{amodal} scene layout from just a single image (\cf{} Fig.~\ref{fig:teaser_new}).
    
    \item We demonstrate that adversarial learning can be used to further enhance the quality of the estimated layouts, specifically when hallucinating large missing chunks of a scene (\cf{} Fig.~\ref{fig:teaser_new}, Sec.~\ref{sec:approach}).
    
    
    \item We evaluate against several state-of-the-art approaches, and outperform all of them by a significant margin on a number of established benchmarks (KITTI-Raw, KITTI-Object, KITTI-Odometry \cite{KITTI}, Argoverse \cite{chang2019argoverse}, \cf{} Sec.~\ref{sec:results}, Table~\ref{table:quantitative:main}).
    
    \item Further, we show that \monolayout{} can also be efficiently trained on datasets that do not contain lidar scans by leveraging recent successes in monocular depth estimation. \cite{godard2018digging} (\cf{} Table~\ref{table:ablation_registration}).
\end{compactenum}

Please refer to the appendix for more results, where we demonstrate that the extracted amodal layouts can suit several higher level tasks, such as (but not limited to) multi-object tracking, trajectory forecasting, etc.





%% file: text/relatedwork.tex
\section{Related Work} \label{relatedwork}



To the best of our knowledge, no published approach has tackled the task of simultaneous road layout (static scene) and traffic participant (dynamic scene) estimation from a single image. However, several recent approaches have addressed the problem of estimating the layout of a road scene, and several other independent approaches have tackled 3D object detection. We summarize the most closely related approaches in this section.

\subsubsection*{\textbf{Road layout estimation}}



Schulter et al. \cite{schulter2018learning} proposed one of the first approaches to estimate an occlusion-reasoned \bev{} road layout from a single color image. They use monocular depth estimation \cite{godard2018digging} as well as semantic segmentation to bootstrap a CNN that predicts occluded road layout. They use priors from OpenStreetMap \cite{OpenStreetMap} to adversarially regularize the estimates. More recently, Wang \etal{} \cite{wang2019parametric} builds on top of \cite{schulter2018learning} to infer parameterized road layouts. Our approach does not require to be bootstrapped using either semantic segmentation or monocular depth estimation, and can be trained end-to-end from color images.

Perhaps the closest approach to ours is MonoOccupancy \cite{lu2019monocular}, which builds a variational autoencoder (VAE) to predict road layout from a given image. They also present results for extracting regions close to roads (eg. sidewalk, terrain, non-free space). However, they reason only about the pixels present in the image, and not beyond occluding obstacles. Also, the bottleneck enforced by that leads to non-sharp, blob-like layouts. On the other hand, \monolayout{} estimates \emph{amodal} scene layouts, reasoning beyond occlusion boundaries. Our approach produces crisp road edges as well as vehicle boundaries, by leveraging adversarial feature learning and sensor fusion to reduce noise in the labeled ground-truth training data.

 \subsubsection*{\textbf{Object detection in bird's eye view}}
 
There exist several approaches to 3D object detection that exclusively use lidar \cite{beltran2018birdnet, yang2018pixor, shi2019pointrcnn}, or a combination of camera and lidar sensors \cite{ku2018joint, chen2017multi, liang2018deep}. However, there are only a handful of approaches that purely use monocular vision for object detection \cite{chen2016monocular, li2019gs3d,mousavian20173d}. Most of these are two stage approaches, comprising a region-proposal stage, and a classification stage. 

Another category of approaches map a monocular image to a \bev{} representation \cite{roddick2018orthographic}, thereby reducing the task of 3D object detection to that of 2D image segmentation. Recently, BirdGAN \cite{srivastava2019learning} leveraged adversarial learning for mapping images to \bev{}, where lidar object detectors such as \cite{beltran2018birdnet} were repurposed for object detection.

Such techniques usually require a pre-processing stage (usually a neural network that maps an image to a \bev{}) after which further processing is applied. On the other hand, we demonstrate that we can achieve significantly higher accuracy by directly mapping from the image space to objects in \bev{}, bypassing the need for a pre-processing stage altogether.

More notably, all the above approaches require a post-processing step that usually involves non-maximum suppression / thresholding to output object detections. \monolayout{} neither requires pre-processing nor post-processing and it directly estimates scene layouts that can be evaluated (or plugged into other task pipelines).

%% file: text/approach.tex
\section{MonoLayout: Monocular Layout Estimation}
\label{sec:approach}

\subsection{Problem Formulation}

In this paper, we address the problem of \emph{amodal} scene layout estimation from a single color image. Formally, given a color image $\mathcal{I}$ captured from an autonomous driving platform, we aim to predict a \bev{} layout of the static and dynamic elements of the scene. Concretely, we wish to estimate the following three quantities. \footnote{\emph{Flat-earth assumption}: For the scope of this paper, we assume that the autonomous vehicle is operating within a bounded geographic area of the size of a typical city, and that all roads in consideration are \emph{somewhat} planar, i.e., no steep/graded roads on mountains.}

\begin{compactenum}
    \item The set of all static scene points $\mathcal{S}$ (typically the road and the sidewalk) on the ground plane (within a rectangular range of length $L$ and width $W$, in front of the camera), \emph{regardless of whether or not they are imaged by the camera}.
    \item The set of all dynamic scene points $\mathcal{D}$ on the ground plane (within the same rectangular range as above) occupied by vehicles, \emph{regardless of whether or not they are imaged by the camera}.
    \item For each point discerned in the above step as being occupied by a vehicle, an instance-specific labeling of which vehicle the point belongs to.
\end{compactenum}

\subsection{MonoLayout}

The problem of amodal scene layout estimation is challenging from a neural networks standpoint in several interesting ways. First, it necessitates that we learn \emph{good} visual representations from images that help in estimating 3D properties of a scene. Second, it requires these representations to reason beyond classic 3D reconstruction; these representations must enable us to \emph{hallucinate} 3D geometries of image regions that are occupied by occluders. Furthermore, the learned representations must implicitly disentangle the static parts of the scene (occupied by road points) from the dynamic objects (eg. parked/moving cars). With these requirements in mind, we design \monolayout{} with the following components.

\subsubsection*{\textbf{Maximum a posteriori estimation}}

We formulate the amodal road layout estimation problem as that of recovering the Maximum a posteriori (MAP) estimate of the distribution of scene statics and dynamics. Given the image $\mathcal{I}$, we wish to recover the posterior $P(\mathcal{S}, \mathcal{D} | \mathcal{I})$, over the domain 
$\Omega \triangleq \{(x, H, z) | \|(x-x_0)\|_1 \leq L; \ \ \|(z-z_0)\|_1 \leq W; (z-z_0) > 0 \}$\footnote{This domain is a rectangular region in \bev{}. $H$ is the height of the camera above the ground.}. 
Note that the static (road) and dynamic (vehicle) marginals are not independent. They are not independent - they exhibit high correlation (vehicles ply on roads). Hence, we introduce an additional conditioning context variable $\mathcal{C}$ that can be purely derived only using the image information $\emph{I}$, such that, $\mathcal{S}$ and $\mathcal{D}$ are conditionally independent given $\mathcal{C}$. We term this conditioning variable as the "shared context" as it necessarily encompasses the information needed to estimate the static and dynamic layout marginals. This allows the posterior to be factorized in the following form.

\begin{equation}
    \begin{split}
        P(\mathcal{S}, \mathcal{D} | \mathcal{I}) & \propto P(\mathcal{S}, \mathcal{D}, \mathcal{C} | \mathcal{I}) \\
        & = \underbrace{P(\mathcal{S} | \mathcal{C}, \mathcal{I})}_{\text{static decoder}} \  \underbrace{P(\mathcal{D} | \mathcal{C}, \mathcal{I})}_{\text{dynamic decoder}} \  \underbrace{P(\mathcal{C} | \mathcal{I})}_{\text{context encoder}}
    \end{split}
    \label{eqn:map-estimation}
\end{equation}

In accordance with the above factorization of the posterior, the architecture of \monolayout{} comprises three subnetworks (\cf{} Fig.~\ref{fig: main_architecture}).

\begin{compactenum}
    \item A \emph{context encoder} which extracts multi-scale feature representations from the input monocular image. This provides a shared context that captures static as well as dynamic scene components for subsequent processing.
    \item An \emph{amodal static scene decoder} which decodes the shared context to produce an amodal layout of the static scene. This model consists of a series of deconvolution and upsampling layers that map the shared context to a static scene bird's eye view.
    \item A \emph{dynamic scene decoder} which is architecturally similar to the road decoder and predicts the vehicle occupancies in bird's eye view.
    \item Two \emph{discriminators}\cite{salimans2016improved, isola2017image} which regularize the predicted static/dynamic layouts by regularizing their distributions to be \emph{similar} to the \emph{true} distribution of plausible road geometries (which can be easily extracted from huge unpaired databases such as OpenStreetMap \cite{OpenStreetMap}) and ground-truth vehicle occupancies. 
\end{compactenum}

\subsubsection*{Feature Extractor}

From the input image, we first extract meaningful image features at multiple scales, using a ResNet-18 encoder (pre-trained on ImageNet\cite{deng2009imagenet}). We finetune this feature extractor in order for it to learn low-level features that help reason about static and dynamic aspects of the scene.

\subsubsection*{Static and dynamic layout decoders}

The static and dynamic layout decoders share an identical architecture. They decode the shared context from the feature extractor by a series of upsampling layers to output a $D \times D$ grid each\footnote{We tried training a single decoder for both the tasks, but found convergence to be hard. We attribute this to the extreme change in output spaces: while roads are large, continuous chunks of space, cars are tiny, sparse chunks spread over the entire gamut of pixels. Instead, we chose to train two decoders over a shared context, which bypasses this discrepancy in output spaces, and results in sharp layout estimates.}.

\subsubsection*{Adversarial Feature Learning}

To better ground the likelihoods $P(\mathcal{S}|\mathcal{C}, \mathcal{I})$, $P(\mathcal{D}|\mathcal{C}, \mathcal{I})$ (\cf Eq.~\ref{eqn:map-estimation}), we introduce adversarial regularizers (discriminators) parameterized by $\theta_S$ and $\theta_D$ respectively. The layouts estimated by the static and dynamic decoders are input to these patch-based discriminators\cite{isola2017image}. The discriminators regularize the distribution of the output layouts (\emph{fake data} distribution, in GAN \cite{GAN} parlance) to match a prior data distribution of \emph{conceivable} scene layouts (\emph{true data} distribution). This prior data distribution is a collection of road snippets from OpenStreetMap \cite{OpenStreetMap}, and rasterized images of vehicles in \bev{}. Instead of training with a paired OSM for each image, we choose to collect a set of diverse OSM maps representing the true data distribution of road layouts in \bev{} and train our discriminators in an unpaired fashion. This mitigates the need to have perfectly aligned OSM views to the current image, making \monolayout{} favourable compared to approaches like \cite{schulter2018learning} that perform an explicit alignment of the OSM before beginning processing.

\subsubsection*{Loss function}

The parameters $\phi, \nu, \psi$ of the context encoder, the amodal static scene decoder, and the dynamic scene decoder respectively are obtained by minimizing the following objective using minibatch stochastic gradient descent.

\begin{equation*}
    \begin{aligned}
    & \underset{\phi, \nu, \psi, \theta_S, \theta_D}{\text{min}} \mathcal{L}_{sup}(\phi, \nu, \psi) + \mathcal{L}_{adv}(\phi, \theta, \psi) + \mathcal{L}_{discr}(\phi, \nu) &\\
    & \mathcal{L}_{sup} =  \sum_{i=1}^N \| \mathcal{S}_{\phi, \nu}(\mathcal{I}^i) - \mathcal{S}^i_{gt} \|^2 + \| \mathcal{D}_{\phi, \psi}(\mathcal{I}^i) - \mathcal{D}^i_{gt} \|^2 & \\
    & \mathcal{L}_{adv}(S, D; \phi, \theta, \psi) = \mathbb{E}_{\theta \sim p_{fake}}\left[(D(\theta_S) - 1)^2\right] &\\
    & +\mathbb{E}_{\theta \sim p_{fake}}\left[(D(\theta_D) - 1)^2\right] &\\
    & \mathcal{L}_{discr}(D; \theta) = \sum_{\theta \in \{\theta_D, \theta_S\}} \mathbb{E}_{\theta \sim p_{true}}\left[(D(\theta) - 1)^2\right] & \\
    & + \mathbb{E}_{\theta \sim p_{fake}}\left[(D(\theta) - 1)^2\right] &
    \end{aligned}
\end{equation*}


Here, $\mathcal{L}_{sup}$ is a supervised ($L2$) error term that penalizes the deviation of the predicted static and dynamic layouts ($\mathcal{S}_{\phi, \nu}(\mathcal{I}^^i)$, $\mathcal{D}_{\phi, \nu}(\mathcal{I}^^i)$) with their corresponding ground-truth values ($\mathcal{S}^i_{gt}$, $\mathcal{D}^i_{gt}$). The adversarial loss $\mathcal{L}_{adv}$ encourages the distribution of layout estimates from the static/dynamic scene decoders ($p_{fake}$) to be close to their true counterparts ($p_{true})$. The discriminator loss $\mathcal{L}_{discr}$ is the discriminator update objective \cite{GAN}.

\begin{table*}[!t]
\begin{center}
\begin{adjustbox}{max width=\linewidth, totalheight=5cm}
\begin{tabular}{c|c|c|c|c|c|c}
\centering
Dataset                         & Method                                        & \multicolumn{3}{c|}{Static Layout Estimation}             & \multicolumn{2}{c}{Vehicle Layout}       \\ \hline
                                &                                               & Road           & Sidewalk    & Road + Sidewalk &                           &        \\ \cline{3-7} 
                                &                                               & mIoU           & mIoU        & occl mIoU       & \multicolumn{1}{c|}{mIoU} & \multicolumn{1}{c}{mAP}   \\ \cline{1-7} 
\multirow{3}{*}{KITTI Raw}      & MonoOccupancy \cite{lu2019monocular}          & $56.16$        &      $18.18$       &    $28.24$             &           -              &  -                        \\
                                & Schulter \etal{} \cite{schulter2018learning}  & $68.89$        &  $30.35$    &   $61.06$             &             -              &  -                        \\
                                & \textbf{MonoLayout-static (Ours)}             & $\mathbf{73.86}$   &  $\mathbf{32.86}$           &    $\mathbf{67.42} $            &         -                 &  -                        \\ \hline
\multirow{2}{*}{KITTI Odometry} & MonoOccupancy \cite{lu2019monocular}          & $64.72$        &  $12.08$           &   $34.87$              &             -             &  -                        \\
                                & \textbf{MonoLayout-static (Ours)}             & $\mathbf{80.08}$   &  $\mathbf{42.66}$                &   $\mathbf{72.46}$         &           -                 &  -                        \\ \hline
\multirow{4}{*}{KITTI Object}   & MonoOccupancy-ext                             &  -             &  -          &  -              & $20.45$                   & $22.59$                    \\
                                & Mono3D \cite{chen2016monocular}               &  -             &  -          &  -              & $17.11$                   & $29.62$                    \\
                                & OFT \cite{roddick2018orthographic}            &  -             &  -          &  -              & $25.24$                   & $34.69$                    \\
                                & \textbf{MonoLayout-dynamic (Ours)}            &  -             &  -          &  -              & $\mathbf{26.08}$              & $\mathbf{40.79}$                 \\ \hline
\multirow{1}{*}{KITTI Tracking} & \textbf{MonoLayout (Ours)}                    & $\mathbf{53.94}$   &  -          &  -              & $\mathbf{24.16}$              & $\mathbf{36.83}$                    \\ \hline
\multirow{2}{*}{Argoverse}      & MonoOccupancy-ext                             & $32.70$        &  -          &  -              & $16.22$                   & $38.66$                    \\
                                & \textbf{MonoLayout (Ours)}                    & $\mathbf{58.33}$   &  -          &  -              & $\mathbf{32.05}$              & $\mathbf{48.31}$ \\        
\end{tabular}
\end{adjustbox}
\end{center}
\caption{\textbf{Quantitative results}: We evaluate the performance of \monolayout{} on several datasets, on amodal scene layout estimation. As there is no existing baseline that simultaneously estimates static (road/sidewalk) as well as dynamic (vehicle) layout, we evaluate under multiple settings. On the KITTI Raw and  KITTI Odometry \cite{KITTI} datasets, we evaluate \monolayout-static. On the KITTI Object \cite{KITTI} dataset, we evaluate \monolayout-dynamic. On the KITTI Tracking \cite{KITTI} and Argoverse \cite{chang2019argoverse} datasets, we evaluate \monolayout, the full architecture that estimates both static and dynamic layouts. We outperform existing approaches by a significant margin, on all metrics.}
\label{table:quantitative:main}
\end{table*}

\subsection{Generating training data: sensor fusion}
\label{sec:sensor-fusion}

Since we aim to recover the amodal scene layout, we are faced with the problem of extracting training labels for even those parts of the scene that are occluded from view. While recent autonomous driving benchmarks provide synchronized lidar scans as well as semantic information for each point in the scan, we propose a sensor fusion approach to generate more robust training labels, as well as to handle scenarios in which direct 3D information (eg. lidar) may not be available.

As such, we use either monocular depth estimation networks (Monodepth2 \cite{godard2018digging}) or raw lidar data and initialize a pointcloud in the camera coordinate frame. Using odometry information over a window of $W$ frames, we aggregate/register the sensor observations over time, to generate a more dense, noise-free pointcloud. Note that, when using monocular depth estimation, we discard depths of points that are more than $5$ meters away from the car, as they are noisy. To compensate for this narrow field of view, we aggregate depth values over a much larger window size (usually $40-50$) frames.

The dense pointcloud is then projected to an occupancy grid in \bev{}. If ground-truth semantic segmentation labels are available, each occupancy grid cell is assigned the label based on a simple majority over the labels of the corresponding points. For the case where ground-truth semantic labels are unavailable, we use a state-of-the-art semantic segmentation network \cite{rota2018place} to segment each frame and aggregate these predictions into the occupancy grid.

For vehicle occupancy estimation though, we rely on ground-truth labels in \bev{}, and train only on datasets that contain such labels\footnote{For a detailed description of the architecture and the training process, we refer the reader to the appendix}.

%% file: text/results.tex
\section{Experiments}
\label{sec:results}

To evaluate \monolayout{}, we conduct experiments over a variety of challenging scenarios and against multiple baselines, for the task of \emph{amodal} scene layout estimation.

\begin{figure*}[!ht]
  \centering
    \begin{adjustbox}{max width=0.8\linewidth, totalheight=7cm}  \includegraphics[width=0.8\linewidth]{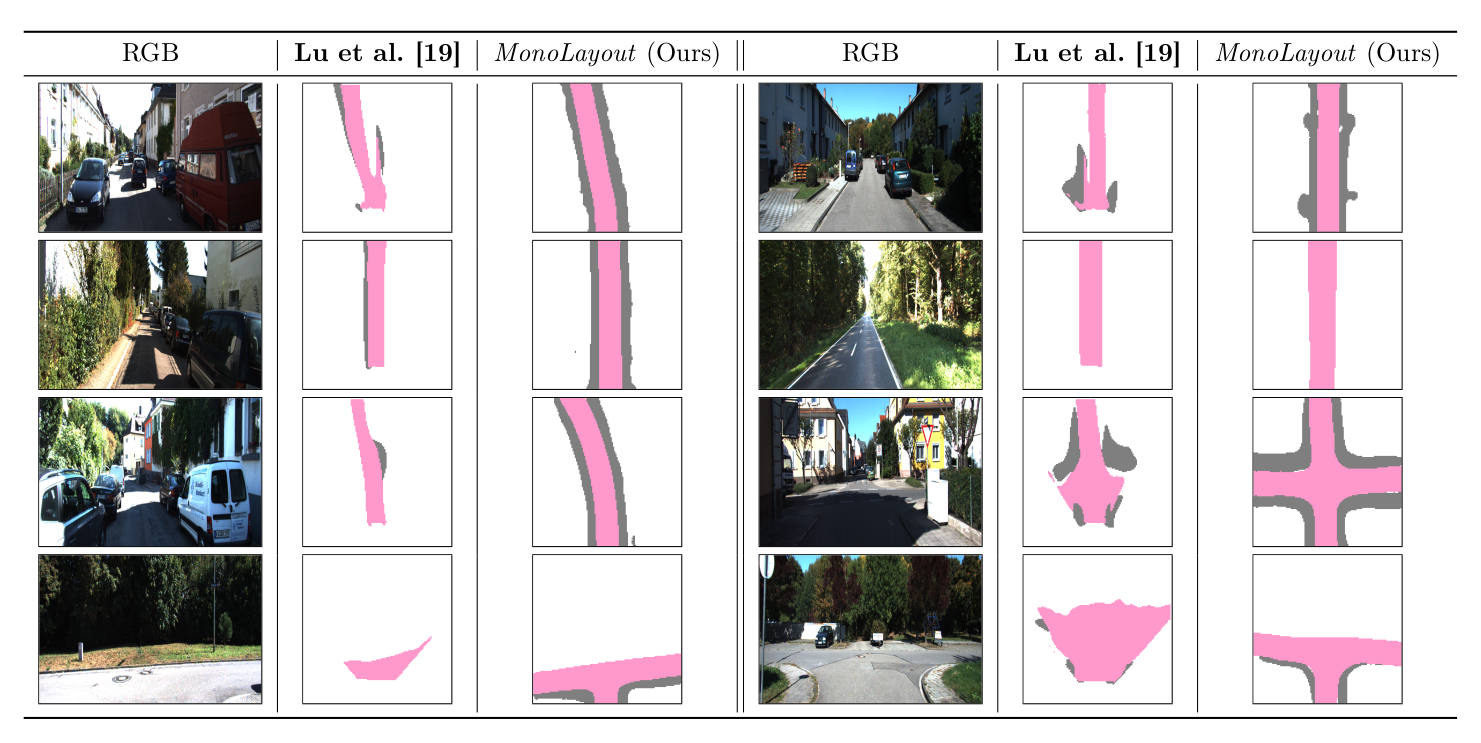}
    \end{adjustbox}
  \caption{\textbf{Static layout estimation}: Observe how \monolayout{} performs amodal completion of the static scene (road shown in {\textcolor{pink}{pink}}, sidewalk shown in {\textcolor{gray}{gray}}. MonoOccupancy \cite{lu2019monocular} fails to reason beyond occluding objects (top row), and does not hallucinate large missing patches (bottom row), while \monolayout{}(Ours) is accurately able to do so. Furthermore, even in cases where there is no occlusion (row 2), \monolayout{}(Ours) generates road layouts of much sharper quality. Row 3 show extremely challenging scenarios where most of the view is blocked by vehicles, and the scenes exhibit high-dynamic range (HDR) and shadows.}
  \label{fig:qualitative_road}
  \vspace{-0.5cm}
\end{figure*}
\subsection{Datasets}

We present our results on two different datasets - KITTI\cite{KITTI} and Argoverse\cite{chang2019argoverse}. The  latter contains a \emph{high-resolution} semantic occupancy grid in \bev, which facilitates the evaluation of amodal scene layout estimation. The KITTI dataset, however, has no such provision. For a semblance of \emph{ground-truth}, we register depth and semantic segmentation of lidar scans in bird's eye view. 

To the best of our knowledge, there is no published prior work that reasons jointly about road and vehicle occupancies. However, there exist approaches for road layout estimation \cite{schulter2018learning, lu2019monocular}, and a separate set of approaches for vehicle detection \cite{roddick2018orthographic,chen2016monocular}. Furthermore, each of these approaches evaluate over different datasets (and in cases, different train/validation splits). To ensure fair comparision with all such approaches, we organize our results into the following categories.

\begin{enumerate}

    \item \textbf{Baseline Comparison:} For a fair comparison with state-of-the-art road layout estimation techniques, we evaluate performance on the KITTI RAW split used in \cite{schulter2018learning} ($10156$ training images, $5074$ validation images). For a fair comparision with state-of-the-art 3D vehicle detection approaches we evaluate performance on the KITTI 3D object detection split of Chen et al.\cite{chen2016monocular} ($3712$ training images, $3769$ validation images).
    
    \item \textbf{Amodal Layout Estimation:} To evaluate layout estimation on both static and dynamic scene attributes (road, vehicles), we use the KITTI Tracking \cite{KITTI} and Argoverse \cite{chang2019argoverse} datasets. We annotate sequences from the KITTI Tracking dataset for evaluation ($5773$ training images, $2235$ validation images). Argoverse provides HD maps as well as vehicle detections in \bev{} ($6723$ training images, $2418$ validation images).
    
    \item \textbf{Temporal sensor fusion for supervision:} We then present results using our data generation approach (\cf{} Sec.~\ref{sec:sensor-fusion}) on the KITTI Odometry dataset. This also uses the dense semantic segmentation labels from the Semantic KITTI dataset \cite{behley2019iccv}.
    
\end{enumerate}

\subsection{Approaches evaluated}

We evaluate the performance of the following approaches.
\begin{itemize}
    \item \emph{Schulter \etal{}}: The static scene layout estimation approach proposed in \cite{schulter2018learning}.
    \item \emph{MonoOccupancy}: The static scene layout estimation approach proposed in \cite{lu2019monocular}.
    \item \emph{Mono3D}: The monocular 3D object detection approach from \cite{chen2016monocular}.
    \item \emph{OFT}: A recent, state-of-the-art monocular \bev{} detector \cite{roddick2018orthographic}.
    \item \emph{MonoOccupancy-ext}: We extend MonoOccupancy \cite{lu2019monocular} to predict vehicle occupancies.
    \item \emph{MonoLayout-static}: A version of \monolayout{} that only predicts static scene layouts.
    \item \emph{MonoLayout-dynamic}: A version of \monolayout{} that only predicts vehicle occupancies.
    \item \monolayout{}: The full architecture, that predicts both static and dynamic scene layouts.
\end{itemize}

\subsection{Static layout estimation (Road)}
\vspace{-0.2cm}

We evaluate \monolayout-static against Schulter \etal{} \cite{schulter2018learning} and MonoOccupancy \cite{lu2019monocular} on the task of static scene (road/sidewalk) layout estimation. Note that Schulter \etal{} assume that the input image is first passed through monocular depth estimation and semantic segmentation networks, while we operate directly on raw pixel intensities. Table~\ref{table:quantitative:main} summarizes the performance of existing road layout estimation approaches (Schulter \etal{} \cite{schulter2018learning}, MonoOccupancy \cite{lu2019monocular}) on the KITTI Raw and KITTI Odometry benchmarks. For KITTI Raw, we follow the exact split used in Schulter \etal{} and retrain MonoOccupancy and \monolayout-static on this train split. Since the manual annotations for semantic classes for KITTI Raw aren't publicly available, we manually annotated sequences with semantic labels (and will make them publicly available).

From Table~\ref{table:quantitative:main} (\cf{} ``KITTI Raw"), we see that \monolayout-static outperforms both MonoOccupancy \cite{lu2019monocular} and Schulter \etal{} \cite{schulter2018learning} by a significant margin. We attribute this to the strong hallucination capabilities of \monolayout-static due to adversarial feature learning (\cf{} Fig.~\ref{fig:qualitative_road}). Although Schulter \etal{} \cite{schulter2018learning} use a discriminator to regularize layout predictions, they seem to suffer from cascading errors due to sequential processing blocks (eg. depth, semantics extraction). MonoOccupancy \cite{lu2019monocular} does not output sharp estimates of road boundaries by virtue of being a variational autoencoder (VAE), as mean-squared error objectives and Gaussian prior assumptions often result in \emph{blurry} generation of samples \cite{lin2019wise}. The hallucination capability is much more evident in the occluded region evaluation, where we see that \monolayout-static improves by more than $10\%$ on prior art.

\begin{figure*}[!ht]
  \centering
    \begin{adjustbox}{max width=0.8\linewidth, totalheight=7cm}  \includegraphics[width=0.8\linewidth]{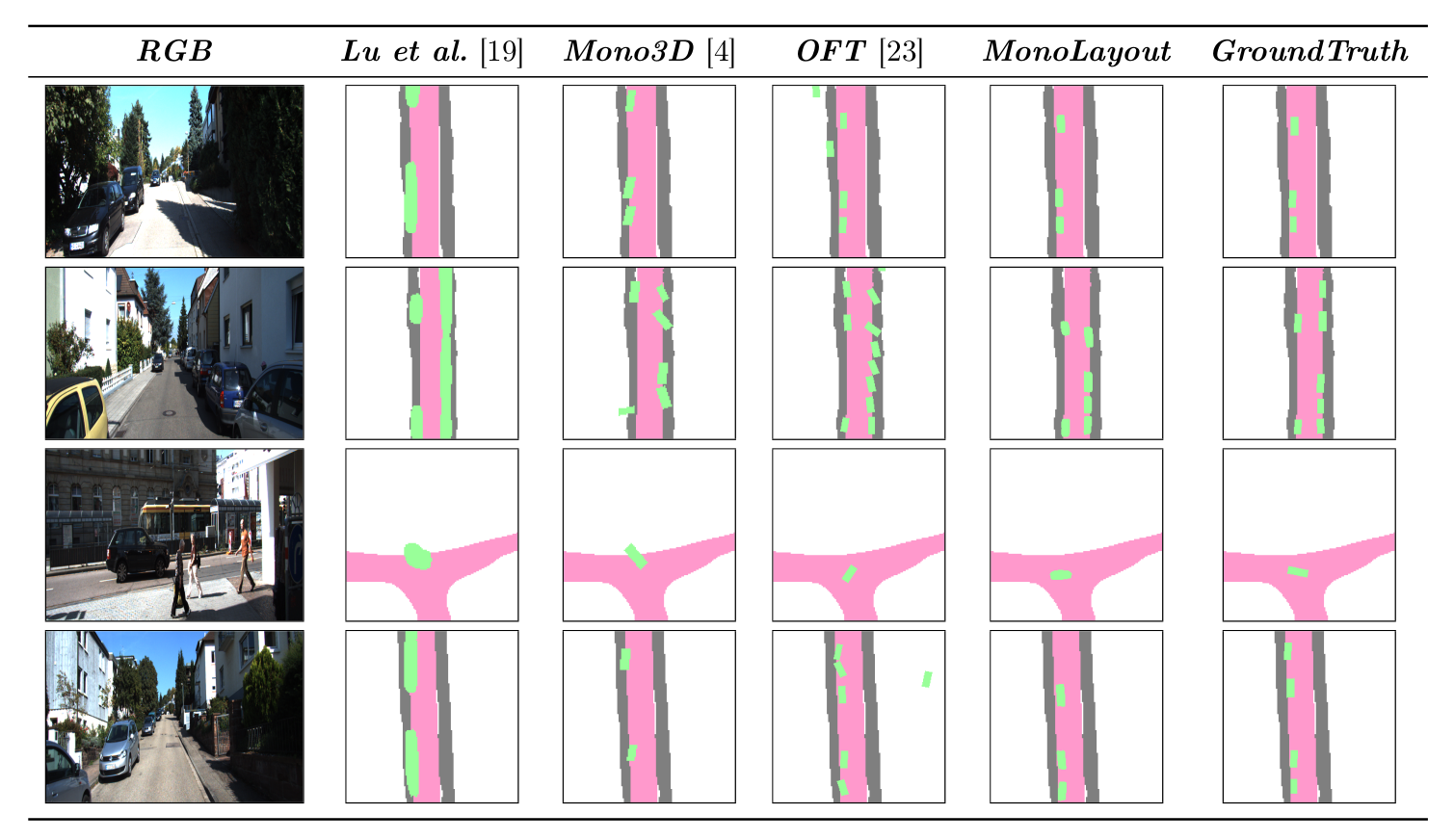}
    \end{adjustbox}
  \caption{\textbf{Dynamic layout estimation}: We show vehicle occupancy estimation results on the KITTI \cite{KITTI} 3D Object detection benchmark. From left to right, the column corresponds to the input image, MonoOccupancy \cite{lu2019monocular}, Mono3D\cite{chen2016monocular}, OFT \cite{roddick2018orthographic}, \monolayout{} (Ours), and ground-truth respectively. While the other approaches miss out on detecting cars (top row), or split a vehicle detection into two (second row), or stray detections off road (third row), \monolayout{} (Ours) produces crisp object boundaries while respecting vehicle and road geometries.}
  \label{fig: qualitative_vehicle}
  \vspace{-0.5cm}
\end{figure*}

\subsection{Dynamic (vehicle) layout estimation}

To evaluate vehicle layout estimation in \bev{}, we first present a comparative analysis on the KITTI Object \cite{KITTI} dataset, for a fair evaluation with respect to prior art. Specifically, we compare against \emph{Orthographic Feature Transform} (OFT \cite{roddick2018orthographic}), the current best monocular object detector in \bev{}. We also evaluate against Mono3D \cite{chen2016monocular}, as a baseline. And, we extend MonoOccupancy \cite{lu2019monocular} to perform vehicle layout estimation, to demonstrate that VAE-style architectures are ill-suited to this purpose. This comparison is presented in Table~\ref{table:quantitative:main} (``KITTI Object'').

We see again that \monolayout-dynamic outperforms prior art on the task of vehicle layout estimation. Note that Mono3D \cite{chen2016monocular} is a two-stage method and requires strictly additional information (semantic and instance segmentation), and OFT \cite{roddick2018orthographic} performs explicit orthographic transfors and is parameter-heavy ($23.5$M parameters) which slows it down considerably ($5$ fps). We make no such assumptions and operate on raw image intensities, yet obtain better performance, and about a $6$x speedup ($32$ fps). Also, MonoOccupany \cite{lu2019monocular} does not perform well on the vehicle occupancy estimation task, as the variational autoencoder-style architecture usually \emph{merges} most of the vehicles into large \emph{blob-like} structures (\cf{} Fig.~\ref{fig: qualitative_vehicle}).

\subsection{Amodal scene layout estimation}

In the previous sections, we presented results individually for the tasks of static (road) and dynamic (vehicle) layout estimation, to facilitate comparision with prior art \emph{on equal footing}. We now present results for \emph{amodal} scene layout estimation (i.e., both static and dynamic scene components) on the Argoverse\cite{chang2019argoverse} dataset. We chose Argoverse \cite{chang2019argoverse} as it readily provides ground truth HD \bev{} maps for both road and vehicle occupancies. We follow the train-validation splits provided by \cite{chang2019argoverse}, and summarize our results in Table~\ref{table:quantitative:main} (``Argoverse''). 

\begin{figure}[!t]
    \centering
    \begin{adjustbox}{max width=\linewidth}
    \includegraphics[]{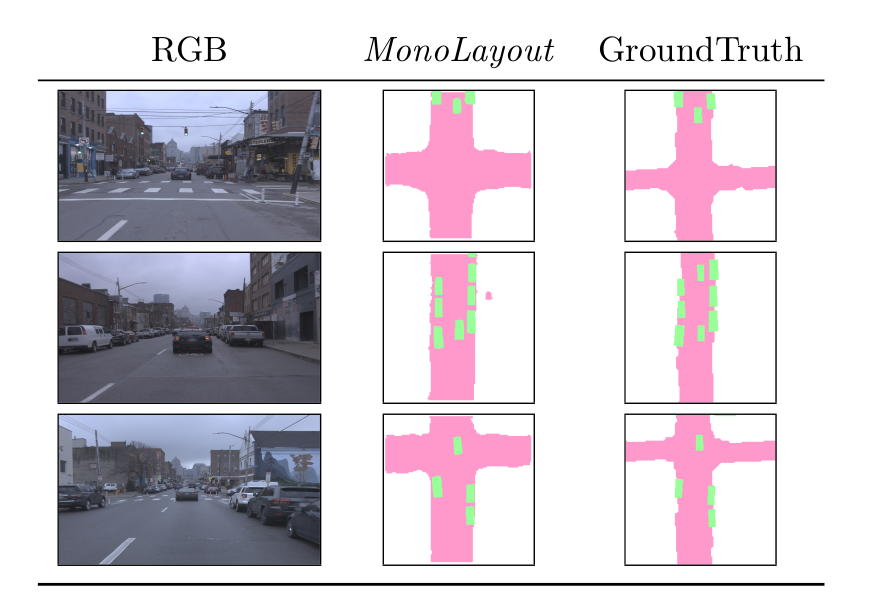}
    \end{adjustbox}
    \caption{\textbf{Amodal scene layout estimation} on the Argoverse \cite{chang2019argoverse} dataset. The dataset comprises multiple challenging scenarios, with low illumination, large number of vehicles. \monolayout{} is accurately able to produce sharp estimates of vehicles and road layouts. (Sidewalks are not predicted here, as they aren't annotated in Argoverse).}
    \label{fig:argo-qualitative}
\end{figure}

We show substantial improvements at nearly $20\%$ on mIoU vis-a-vis the next closest baseline \cite{lu2019monocular}. This is a demonstration of the fact that, even when perfect ground-truth is available, approaches such as MonoOccupancy \cite{lu2019monocular} fail to reach the levels of performance as that of \monolayout{}. We attribute this to the \emph{shared context} that encapsulates a rich feature collection to facilitate both static and dynamic layout estimation. Note that, for the Argoverse\cite{chang2019argoverse} dataset we \emph{do not} train our methods on the ground-truth HD maps, because such fine maps aren't typically available for all autonomous driving solutions. Instead, we train our methods using a semblance of ground-truth (generated by the process described in Sec~\ref{sec:sensor-fusion}), and use the HD maps only for evaluation. This validates our claim that our model, despite being trained using noisy ground estimates by leveraging sensor fusion, is still able to hallucinate and complete the occluded parts of scenes correctly as shown in Fig \ref{fig:argo-qualitative}.

\begin{figure}[!t]
    \centering
    \begin{adjustbox}{max width=\linewidth}
    \includegraphics[]{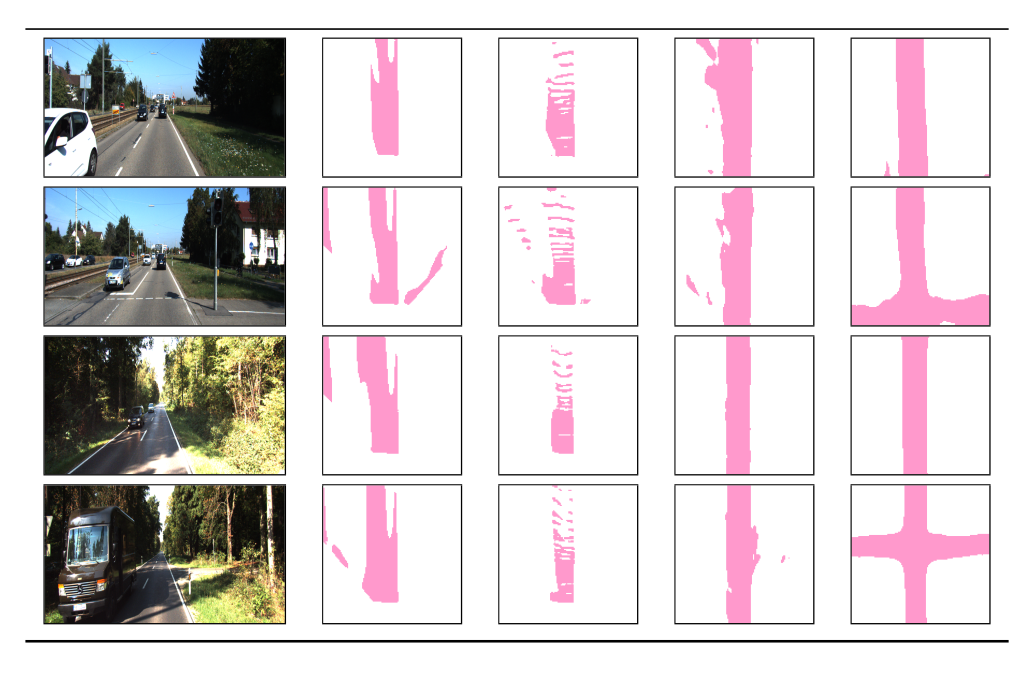}
    \end{adjustbox}
    \caption{\textbf{Impact of sensor fusion}: \emph{Col 1}: Input images. \emph{Col 2}: Using per-frame monocular depth, \emph{Col 3}: Using per-frame lidar depth, \emph{Col 4}: Sensor-fused monocular depth, \emph{Col 5}: Sensor-fused lidar depth. (Refer to Table~\ref{table:ablation_registration} and Sec.~\ref{sec:ablation-monodepth} for more details)}
    \label{fig:sensor-fusion-qualitative}
\end{figure}

\subsection{Ablation Studies}

\subsubsection*{Using monocular depth as opposed to lidar}
\label{sec:ablation-monodepth}

While most of the earlier results focused on the scenario where explicit lidar annotations were available, we turn to the more interesting case where the dataset only comprises monocular images. As described in Sec~\ref{sec:sensor-fusion}, we use monocular depth estimation (MonoDepth2 \cite{godard2018digging}) and aggregate/register depthmaps over time, to provide training signal. In Table~\ref{table:ablation_registration}, we analyze the impact of the availability of lidar data on the performance of amodal scene layout estimation.

We train MonoOccupancy \cite{lu2019monocular} and \monolayout-static on the KITTI Raw dataset, using monocular depth estimation-based ground-truth, as well as lidar-based ground-truth. While lidar based variants perform better (as is to be expected), we see that self-supervised monocular depth estimation results in reasonable performance too. Surprisingly, for the per-frame case (i.e., no sensor fusion), monocular depth based supervision seems to fare better. Under similar conditions of supervision, we find that \monolayout-static outperforms MonoOccupancy \cite{lu2019monocular}.

\subsubsection*{Impact of sensor fusion}
\label{sec:ablation-registration}

The sensor fusion technique described in Sec~\ref{sec:sensor-fusion} greatly enhances the accuracy of static layout estimation. Aggregating sensor observations over time equips us with more comprehensive, and noise-free maps. Table~\ref{table:ablation_registration} presents results for an analysis of the performance benefits obtained due to temporal sensor fusion.

\setlength\extrarowheight{5pt}
\begin{table}[!hbt]
	\centering
    
    \begin{adjustbox}{max width=\linewidth}
		\begin{tabular}{c|c|c}
		\hline
		
		\textbf{Supervision} & \textit{MonoOccupancy-ext} & \monolayout-static (Ours) \\
        
        \hline
        Per-frame monocular depth & $56.16$ & $58.87$ \\
        Sensor-fused monocular depth & $64.81$ & $66.71$ \\
        \hline
        Per-frame lidar & $44.29$ & $48.29$ \\
        Sensor-fused lidar & $\mb{71.67}$ & $\mb{73.86}$ \\
        \hline  

	    \end{tabular}
    \end{adjustbox}
    \caption{\textbf{Monocular depth}: If lidar data is unavailable, we leverage self-supervised monocular depth estimation to generate training data for \monolayout{}-static and achieve reasonable static layout estimation (rows 1-2). Although performance is inferior to the case when lidar is available (rows 3-4), this is not unexpected. \textbf{Sensor fusion}: Regardless of the modality of depth information, sensor-fusing depth estimates over a window of $40$ frames dramatically improves performance (row 2, row 4).}
    \label{table:ablation_registration}
\end{table}

\subsubsection*{Impact of adversarial learning}

\begin{figure}[!t]
    \centering
    \begin{adjustbox}{max width=\linewidth}
    \includegraphics[]{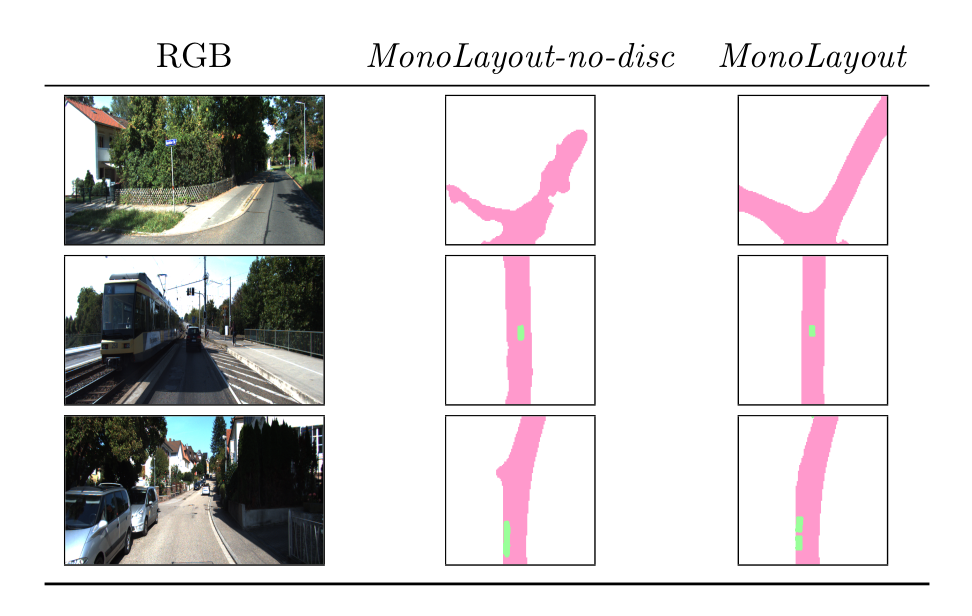}
    \end{adjustbox}
    \caption{\textbf{Effect of adversarial learning}: As can be clearly seen here, the discriminators help enhance both the static (road) layout estimation (top and middle rows), as well as produce sharper vehicle boundaries (bottom row). While this translates to performance gains in static layout estimation (\cf{} Table~\ref{table:ablation_discriminator}), the gains in dynamic layout estimation are more cosmetic in nature.}
    
    \label{fig:discriminator-ablation}
\end{figure}

With the discriminators, we not only improve qualitatively (sharper/realistic samples) (\cf{} Fig~\ref{fig:discriminator-ablation}), but also gain significant performance boosts. This is even more pronounced on Argoverse \cite{chang2019argoverse}, as shown in Table~\ref{table:ablation_discriminator}. In case of vehicle occupancy estimation, while using a discriminator does not translate to quantitative performance gains, it often results in qualitatively sharper, aesthetic estimates as seen in Fig~\ref{fig:discriminator-ablation}.
\vspace{-0.3cm}

\setlength\extrarowheight{15pt}
\begin{table}[!h]
	\centering
    
    \begin{adjustbox}{max width=\linewidth}
	\begin{tabular}{c|c|c|c|c|c|c}
    \textbf{Dataset}       & \multicolumn{3}{c|}{\textbf{\emph{MonoLayout-no-disc}}}                          & \multicolumn{3}{c}{\textbf{\emph{MonoLayout}}}                                  \\ \hline
    \multicolumn{1}{c|}{} & \textbf{Road} & \textbf{Vehicle (mIoU)} & \textbf{Vehicle (mAP)} & \textbf{Road} & \textbf{Vehicle (mIoU)} & \textbf{Vehicle (mAP)} \\ \hline
    KITTI Raw              & $70.95$         & -                       & -                      & $\mb{73.86}$         & -                       & -                      \\
    KITTI Object           & -             & $\mb{26.25}$                   &        $37.66$                & -             & $25.47$                   &   $\mb{41.52}$                     \\
    Argoverse              & $51.66$         & $\mb{32.84}$                   &           $44.07$             & $\mb{58.33}$         & $32.06$                   &  $\mb{48.31}$                     \\ \hline
    \end{tabular}
    \end{adjustbox}
    \caption{\textbf{Effect of discriminator}: Adding a discriminator clearly translates to an accuracy boost in static (road) layout estimation. For vehicle occupancy estimation, while a quantitative boost is not perceived, the generated layouts are sharper, and aesthetic as opposed to when not using the discriminator (\cf{} Fig.~\ref{fig:discriminator-ablation})}
        
    \label{table:ablation_discriminator}
\end{table}

\subsubsection*{Timing analysis}
\vspace{-0.2cm}
 
We also show the computation test time of our method as compared to other baslines in Table \ref{table:computation_time_table}. Unlike Schulter \etal{} \cite{schulter2018learning}, our network does not require discriminator to be used during inference time. It achieves real time inference rate of approx. $32$ Hz for an input size $3 \times 512 \times 512$ and an output size $2 \times 128 \times 128$ on an NVIDIA GeForce GTX 1080Ti GPU.
Note that in \monolayout{} the static and dynamic decoders are executed in parallel, maintining comparable runtime.
\monolayout{} is an order of magnitude faster than previous methods, making it more attractive for on-road implementations.
\vspace{-0.2cm}

\setlength\extrarowheight{7pt}
\begin{table}[!hbt]
	\centering
	
    \begin{adjustbox}{max width=0.6\linewidth}
    \begin{tabular}{|c|c|c|}
		\hline
        \textbf{Method} & \textbf{Parameters} & \textbf{Computation Time} \\
        \hline Mono3D\cite{chen2016monocular} & $>> 20$ M & $0.24$ fps \\
        \hline
        OFT\cite{roddick2018orthographic} & $23.5$ M & $< 5$ fps \\
        \hline
        MonoOccupancy\cite{lu2019monocular} & $27.5$ M & $15$ fps \\
        \hline
        Schulter et al.\cite{schulter2018learning} & $>> 20$ M & $< 3$ fps \\
        \hline
        \monolayout{} (Ours) & $19.6$M & $32$ fps \\
        \hline
	\end{tabular}
    \end{adjustbox}
    \caption{A comparative study of infrence time of various methods. \monolayout{} is about 2x faster and significantly more accurate compared to prior art. (\cf{} Table~\ref{table:quantitative:main}).}
        
    \label{table:computation_time_table}
    \vspace{-0.4cm}
\end{table}

%% file: text/conclusions.tex
\section{Discussion and conclusions}
\label{sec:conclusions}

This paper proposed \monolayout{}, a versatile deep network architecture capable of estimating the amodal layout of a complex urban driving scene in real-time. 
In the appendix, we show several additional results, including extended ablations, and applications to multi-object tracking and trajectory forecasting.
A promising avenue for future research is the generalization of \monolayout{} to unseen scenarios, as well as incorporating temporal information to improve performance.

%% file: text/acknowledgements.tex
\section*{Acknowledgements}

We thank Nitesh Gupta, Aryan Sakaria, Chinmay Shirore, Ritwik Agarwal and Sachin Sharma for manually annotating several sequences from the KITTI RAW\cite{KITTI} dataset.

%% file: text/appendices.tex
    
    

\section{Implementation Details}

In this section, we describe the network architecture and training procedure in greater detail.

\subsection{Network Architecture}

\monolayout{} comprises the following four blocks: a feature extractor, a static layout decoder, dynamic layout decoder, and two discriminators.

\subsubsection*{Feature extractor}

Our feature extractor is built on top of a ResNet-18 encoder\footnote{We also tried other feature extraction / image-image architectures such as UNet and ENet, but found them to be far inferior in practice.}. The network usually takes in RGB images of size $3 \times 512 \times 512$ as input, and produces a $512 \times 32 \times 32$ feature map as output. In particular, we use the ResNet-18 architecture without bottleneck layers. (Bottleneck layers are absent, to ensure fair comparision with OFT~\cite{roddick2018orthographic}). This extracted feature map is what we refer to as the \emph{shared context}.

\subsubsection*{Layout Decoders}

We use two parallel decoders with identical architectures to estimate the static and dynamic layouts. The decoders consists $2$ convolution layers and take as input the $512 \times 32 \times 32$ shared context. The first convolution block maps this shared context to a $128 \times 16 \times 16$ feature map, and the second convolution block maps the output of the first block to another $128 \times 8 \times 8$ feature map.

At this point, $4$ deconvolution (transposed convolution) blocks are applied on top of this feature map. Each block increases spatial resolution by a factor of $2$, and decreases the number of channels to $64, 32, 16$, and $O$ respectively, where $O$ is the number of channels in the output feature map ($O \in \{1, 2\}$ for the static layout decoder, and $O = 1$ for the dynamic layout decoder).
This results in an output feature map of size $O \times 128 \times 128$. We also apply a spatial dropout (ratio of $0.4$) to the penultimate layer, to impose stochastic regularization. The output $128 \times 128$ grid corresponds to a rectangular region of area $40 m \times 40 m$ on the ground plane.

\subsubsection*{Discriminators}

The discriminator architecture is inspired by Pix2Pix~\cite{pix2pix}. We found the patch based regularization in Pix2Pix to be much better than a standard DC-GAN~\cite{dcgan}. So, we use patch-level discriminators that contain four convolution layers (kernel size $3 \times 3$, stride $2$), that outpus an $8 \times 8$ feature map. This feature map is passed through a $tanh$ nonlinearity and used for patch discrimination.

\setlength\extrarowheight{5pt}
\begin{table*}[!hbt]
	\centering
	
    \begin{adjustbox}{max width=0.9\linewidth}
    \begin{tabular}{|c|c|c|c|c|}
		\hline
        \textbf{Method} & \textbf{Vehicle (mAP)} & \textbf{Vehicle (mIoU)} & \textbf{Road (mIou)} & \textbf{Frame Rate} \\
        \hline
        ENet + Pseudo lidar input(Monodepth2) & $0.37$ & $0.24$ & $0.62$ & $ 12.34$ fps \\
        \hline
        PointRCNN + Pseudo lidar input(Monodepth2) & $\mathbf{0.43}$ & $0.26$ & - & $5.76$ fps \\
        \hline
        \monolayout{} (Ours) & $0.41$ & $\mathbf{0.26}$ & $\mathbf{0.80}$ & $\mathbf{32}$ fps\\
        \hline
        \hline
        AVOD +  Pseudo lidar input(PSMNet) (\textbf{Stereo}) & $0.59$ & $0.43$ & - & $ < 1.85$ fps\\
        \hline
	\end{tabular}
    \end{adjustbox}
    \caption{\textbf{Comparision with Pseudo-lidar \cite{wang2019pseudo}}: We also evaluate \monolayout{} against several variants of pseudo-lidar \cite{wang2019pseudo} approaches. While the usage of increasingly heavy processing blocks for pseudo-lidar variants improves accuracy, it drastically increases computation time. On the other hand, \monolayout{} offers---by far---the best mix of accuracy and runtime for real-time applications. Also note that the comparision with AVOD + Pseudo-lidar (PSMNet) is unfair, since it uses stereo disparities.}
        
    \label{table:pseudo_lidar_table}
\end{table*}

\subsection{Training details}

We train \monolayout{} with a batch size of $16$ for $200$ epochs using the Adam~\cite{Adam} optimizer with initial learning rate $5e-5$. The input image is reshaped to $3 \times 512 \times512$ and further augmented to make our model more robust. Some of the augmentation techniques we use include random horizontal flipping and color jittering.

\subsection{Metrics used}

\subsubsection*{Road layout estimation}
To evaluate estimated road layouts, we use intersection-over-union (IoU) as our primary metric. We split IoU evaluation into two parts and measure IoU for the entire static scene, as well as IoU for occluded regions (i.e., regions of the road that are occluded in the image and were hallucinated by \monolayout{}).

\subsubsection*{Vehicle occupancy estimation}
While most approaches to vehicle detection evaluate only mean Average Precision (mAP), it has been shown to be a grossly inaccurate measure of how tight a bounding box is \cite{prod}. We hence adopt mean Intersection-over-Union (mIoU) as our primary basis of evaluation. To ensure a fair comparision with prior art, we also report mAP.\footnote{We outperform existing methods under both these metrics}. (Note that, since we are evaluating object detection in \bev{}, and not in 3D, we use the mIoU and mAP, as is common practice \cite{KITTI,chen2016monocular}). This choice of metrics is based on the fact that \monolayout{} is not an object ``\emph{detection}" approach; it is rather an object occupancy estimation approach, which calls for mIoU and mAP evaluation. We extend the other object ``detection" approaches (such as pseudo-lidar based approaches \cite{shi2019pointrcnn, you2019pseudo, roddick2018orthographic}) to the occupancy estimation setting, for a fair comparision.

\section{Timing analysis}
 
We also show the computation test time of our method as compared to similar methods in Table \ref{table:computation_time_table}. Our network does not require discriminator to be used during inference time. It achieves real time inference rate of approx. $30$ Hz for an input image with a resolution of 512 * 512 pixels and an output map with 128 * 128 occupancy grid cells using a Nvidia GeForce GTX 1080Ti GPU. The code for \cite{schulter2018learning } is not publicly available, and the computation time is based on the PSMNet \cite{psmnet} backbone they use. Here again the proposed method is almost an order faster than previous methods making it more attractive for on-road implementations.

\setlength\extrarowheight{5pt}
\begin{table}[!hbt]
	\centering
	
    \begin{adjustbox}{max width=\linewidth}
    \begin{tabular}{|c|c|c|}
		\hline
        \textbf{Method} & \textbf{Parameters} & \textbf{Computation Time} \\
        \hline
        OFT\cite{roddick2018orthographic} & $24.5$ M & $2$ fps \\
        \hline
        MonoOccupancy\cite{lu2019monocular} & $27.5$ M & $15$ fps \\
        \hline
        Schulter et al.\cite{schulter2018learning} & $>> 20$ M & $< 3$ fps \\
        \hline
        \monolayout{} (Ours) & 19.6M & 32 fps \\
        \hline
	\end{tabular}
    \end{adjustbox}
    \caption{A comparative study of test computation time on NVIDIA GeForce GTX 1080Ti GPU for different methods on the images of KITTI\cite{KITTI} RAW dataset.}
        
    \label{table:computation_time_table}
\end{table}

\section{Comparision with pseudo-lidar}

There is another recent set of approaches to object detection in \bev{}---\emph{pseudo-lidar} approaches \cite{wang2019pseudo}. At the core of these approaches lies the idea that, since lidar object detection works exceedingly well, monocular images can be mapped to (\emph{pseudo}) lidar-like maps in \bev{}, and object detecion networks tailored to lidar \bev{} maps can readily be applied to this setting. Such approaches are primarily geared towards detecting objects in lidar-like maps.

\monolayout{}, on the other hand, intends to estimate an \emph{amodal} scene layout, and to do so, it must reason not only about vehicles, but also about the static scene layout. Table~\ref{table:pseudo_lidar_table} compares \monolayout{} with a set of \emph{pseudo-lidar} approaches, in terms of vehicle occupancy estimation and road layout estimation. Specifically, we evaluate the following pesudo-lidar based methods.
\begin{enumerate}
    \item \emph{ENet + Pseudo-lidar input (Monodepth2)}: Uses an ENet~\cite{enet}-style encoder-decoder architecture that uses Monodepth2~\cite{godard2018digging} to get monocular depth estimates.
    \item \emph{PointRCNN + Pseudo-lidar input (Monodepth2)}: Uses a PointRCNN~\cite{shi2019pointrcnn} architecture (a two-stage object detector comprising a region proposal network, and a classification network) to detect vehicles in \bev{}.
    \item \emph{AVOD + Pseudo-lidar input (PSMNet)}: A \textbf{stereo, supervised} method. Uses the aggregated view object detector AVOD~\cite{avod} and pseudo-lidar input computed from a disparity estimation network (PSMNet~\cite{psmnet}).
\end{enumerate}

The comparision is shown in Table~\ref{table:pseudo_lidar_table}. The PointRCNN~\cite{shi2019pointrcnn} and AVOD~\cite{ku2018joint} are tailored specifically for object detection, and hence \emph{cannot be repurposed to estimate road layouts}. However, the ENet~\cite{enet} architecture can, and we trained it for the task of road layout estimation. We observe that, among all approaches, \monolayout{} is the fastest (about an order of magnitude speedup over pseudo-lidar methods). Furthermore, the accuracy is competitive, if not greater, compared to pseudo-lidar based approaches. 

We also evaluate against a stereo pseudo-lidar baseline (AVOD + pseudolidar PSMNet~\cite{psmnet}). By virtue of using stereo images, and being supervised on the KITTI Object dataset \footnote{Monodepth2~\cite{godard2018digging} is unsupervised, and has not been finetuned on the KITTI Object dataset}, achieves superior performance. However, the comparision is unfair, and is provided only for a reference, to enable progress in amodal layout estimation from a monocular camera.

Another shortcoming of pseudolidar-style approaches is that, it is not possible to learn \emph{visual} (i.e., image intensity based) features that are extremely useful in road layout estimation).

\section{Application: Trajectory forecasting}

One of the use-cases of \monolayout{} is to forecast future trajectories from the estimated \emph{amodal} scene layout. 

We demonstrate accurate trajectory forecasting performance by training a Convolutional LSTM that operates over the estimated layouts from \monolayout{}. Specifically, we adopt an encoder-decoder structure similar to ENet~\cite{enet}, but add a convolutional LSTM between the encoder and the decoder. We also add a convolutional LSTM over each of the skip connections present in ENet.

\begin{figure*}[!t]
    \centering
    \begin{adjustbox}{max width=\textwidth}
    \includegraphics[]{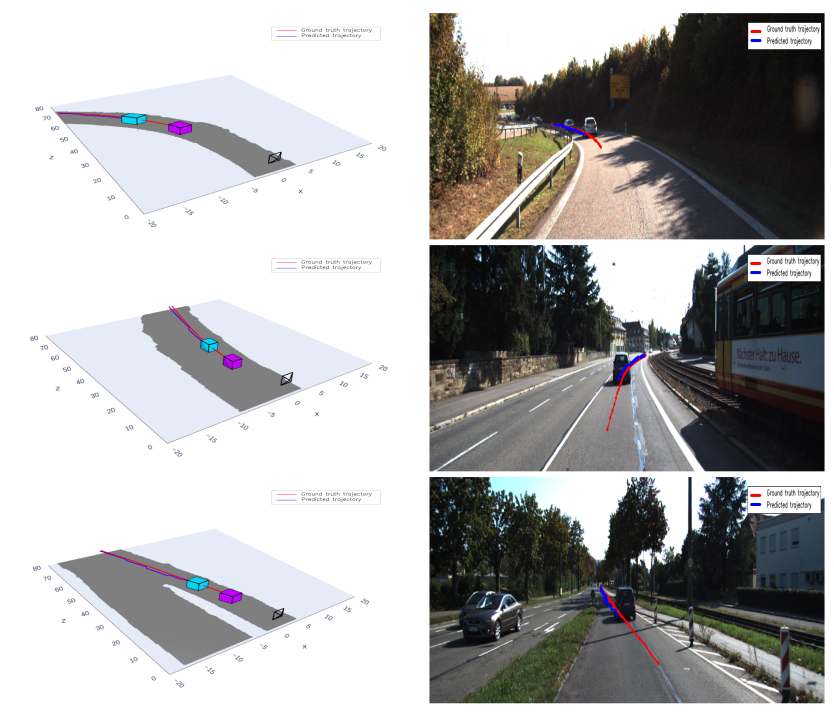}
    \end{adjustbox}
    \caption{\textbf{Trajectory forecasting}: \monolayout-\emph{forecast} accurately estimates future trajectories of moving vehicles. (\emph{Left}): In each figure, the magenta cuboid shows the initial position of the vehicle. \monolayout-\emph{forecast} is pre-conditioned for 1 seconds, by observing the vehicle, at which point (cyan cuboid) it starts forecasting future trajectories (shown in blue). The ground-truth trajectory is shown in red, for comparision. (\emph{Right}): Trajectories visualized in image space. Notice how \monolayout-\emph{forecast} is able to forecast trajectories accurately despite the presence of moving obstacles (top row), turns (middle row), and merging traffic (bottom row).}
    \label{fig:trajectory-forecasting}
\end{figure*}

We \emph{pre-condition} the trajectory forecasting network for $1$ second, by feeding it images, and then predict future trajectories for the next $3$ seconds. Note that, when predicting future trajectories, no images are fed to the forecasting network. Rather, the network operates in an autoregressive manner, by producing an output trajectory estimate, and using this estimate as the subsequent input\emph{We also tried predicting static scene layouts, to \emph{forecast} future static scenes, but without any success, owing to the high variability in static scene layouts}. The resultant model, called \monolayout-\emph{forecast}, works in real-time, and accurately forecasts the future trajectory of a moving vehicle, as shown in Fig.~\ref{fig:trajectory-forecasting}. As with the layout estimation task, the forecast trajectories are confined to a $128 \times 128$ grid, equivalent to a $40 m \times 40 m$ square area in front of the ego-vehicle.

\section{Application: Multi-object tracking}
Further, we propose an extension to generate accurate \emph{tracks} of multiple moving vehicles by leveraging the vehicle occupancy estimation capabilities of \monolayout{}. We also construct a strong baseline multi-object tracker using the open-source implementation from IoU-Tracker~\cite{ioutracker} as a reference. We term this baseline \emph{BEVTracker}. Specifically, we use disparity estimates from stereo cameras, semantic and instance segmentation labels, to segment and identify unique cars in \bev{}. We then run IoU-Tracker~\cite{ioutracker} on these estimates.

We demonstrate in Table~\ref{table:tracking_table} that \monolayout{} outperforms BEVTracker~\cite{ioutracker}, without access to any such specialized information (disparity, semantics, instances). Instead, we run a traditional OpenCV blob detector on \monolayout{} vehicle occupancy estimates, and use the estimated instances to obtain the coordinates of the center of the vehicle. We then use a \emph{maximum-overlap} data association strategy across time, using intersection-over-union to measure overlap. We run our approach on all $24$ sequences of the KITTI Tracking (train) benchmark, and present the results in Table~\ref{table:tracking_table}.

\setlength\extrarowheight{5pt}
\begin{table*}[!hbt]
	\centering
    
    \begin{adjustbox}{max width=\linewidth}
		\begin{tabular}{|c|c|c|c|}
		\hline
		\textbf{Method} & \textbf{Mean Error in Z (m)} & \textbf{Mean Error in X (m)} & \textbf{Mean L2 error (m)}\\
        \hline
        \textbf{BEVTracker} (Open-source \cite{ioutracker}, enhanced with stereo, semantics, etc.) & $1.08$ & $0.51$ & $1.27$ \\
        \hline
        \textbf{\monolayout{} (Ours)} & $\mathbf{0.23}$ & $\mathbf{0.47}$ & $\mathbf{0.58}$ \\
        \hline

	    \end{tabular}
    \end{adjustbox}
    \caption{\textbf{Multi-object tracking performance}: We show the preformance of \monolayout{} on a multi-object tracking task, in \bev{}. The comparision is unfair as \emph{BEVTracker} uses strictly more information (disparities, semantic segmentation, instance segmentation). However, \monolayout{} does not employ any such privileged information, and as such uses OpenCV blob detection to identify instances, and a \emph{maximum-IoU-overlap} data association framework. These estimtaes are obtained over $24$ sequences of the KITTI tracking benchmark. (Caveat: We only evaluate tracking accuracies for cars within a $40 m \times 40 m$ square region around the ego-vehicle, as \monolayout{} estimates are confined to this region).}
    \label{table:tracking_table}
\end{table*}    

\section{More qualitative results}

\begin{figure*}[!t]
    \centering
    \begin{adjustbox}{max width=\textwidth}
    \includegraphics[]{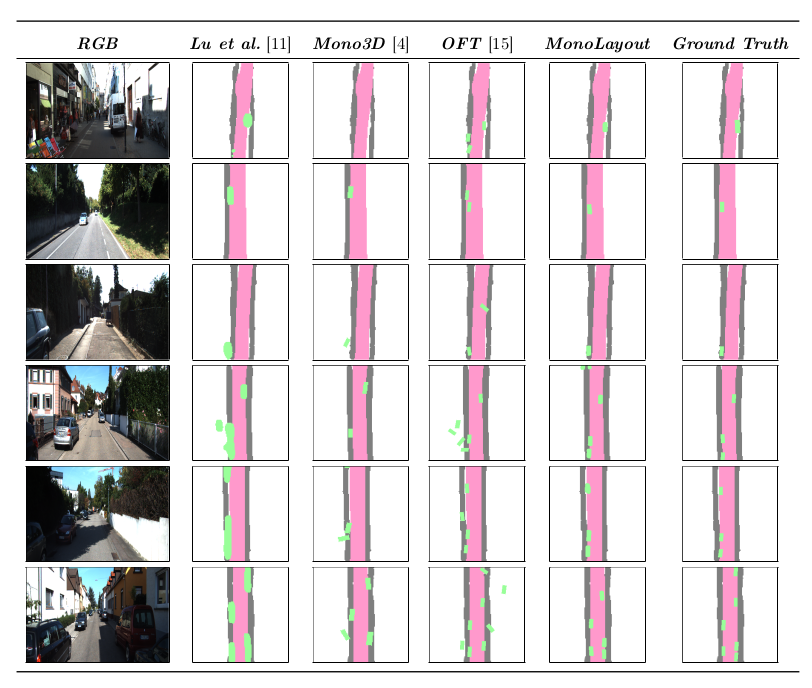}
    \end{adjustbox}
    \caption{\footnotesize{\textbf{Dynamic layout estimation on KITTI \cite{KITTI}}: Additional qualitative results for dynamic scene layout estimation on the KITTI~\cite{KITTI} dataset. From left to right, the column corresponds to the input image, MonoOccupancy \cite{lu2019monocular}, Mono3D\cite{chen2016monocular}, OFT \cite{roddick2018orthographic}, \monolayout{} (Ours), and ground-truth respectively. \monolayout{} (Ours) produces crisp object boundaries while respecting vehicle and road geometries.}}
    \label{fig:dynamic-kitti}.
\end{figure*}

\begin{figure*}[!t]
    \centering
    \begin{adjustbox}{max width=0.5\textwidth}
    \includegraphics[]{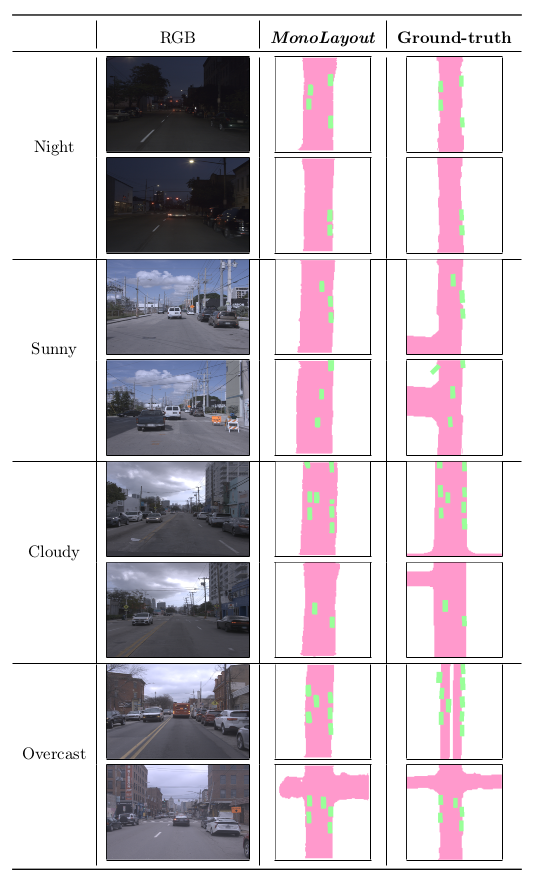}
    \end{adjustbox}
    \caption{\textbf{Qualitative results on Argoverse}: Additional qualitative results on the Argoverse~\cite{chang2019argoverse} dataset (road shown in {\textcolor{pink}{pink}}, vehicles shown in {\textcolor{green}{green}}. \monolayout{} (center column) uses both static and dynamic layout discriminators and produces sharp estimates, and is robust to varying weather conditions, high dynamic range (HDR), and shadows.}
    
    \label{fig:qualitative-weather}
\end{figure*}

Additional qualitative results of joint static and dynamic scene layout estimation are presented in Fig.~\ref{fig:dynamic-kitti} and Fig.~\ref{fig:qualitative-weather}.

\begin{figure*}[!t]
    \centering
    \begin{adjustbox}{max width=\textwidth}
    \includegraphics[]{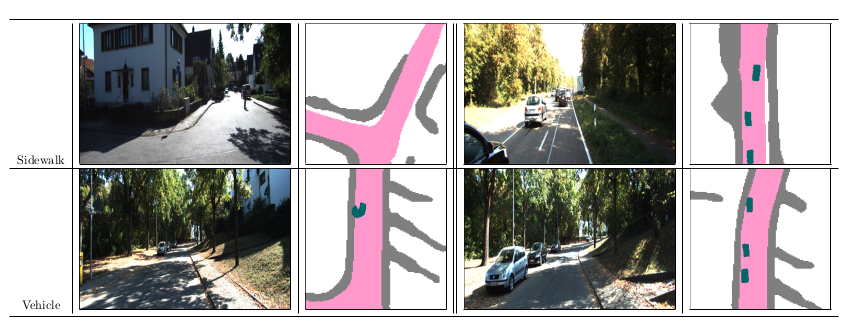}
    \end{adjustbox}
    \caption{\textbf{Failure cases}: This figure highlights a few cases in which \monolayout{} produces incorrect layout estimates. Adverse lighting conditions and sharp turns, in some cases effect sidewalk estimation accuracy(Row 1). Also, multiple near-by vehicles in an image get merged into a single estimate, at times. (As shown in Row 2, only when the ego-vehicle gets close to the two cars parked in close vicinity to each other, the model is able to distinguish the two cars.) }
    \label{fig:failure-cases}
\end{figure*}

\section{Shortcomings}

Despite outperforming several state-of-the-art approaches and achieving real-time performance, \monolayout{} also sufferes a few shortcomings. In this section, we discuss some failure cases of \monolayout{} and also some negative results. Please watch the supplementary video for additional results.

\subsection{Failure cases}

Fig.~\ref{fig:failure-cases} shows a few scenarios in which \monolayout{} fails to produce accurate layout estimates. Recall that \monolayout{} uses adversarial feature learning to estimate plausible road layouts. We leverage OpenStreetMap~\cite{OpenStreetMap} to randomly extract road patches to use as the true data distribution. However, no such data is available for \emph{sidewalks}, and this results in a few artifacts.

As shown in the bottom row of Fig.~\ref{fig:failure-cases}, high-dynamic range and shadows coerce the network into predicting \emph{protrusions} along these directions. Also, as shown in the top row, sometimes, sidewalk predictions are not coherent with road predictions. In the bottom row, we show failure cases for vehicle occupancy estimation.

\subsection{Negative results}

Produced below is a list of experiments that the authors tried, but were unsuccessful. We hope this will expedite progress in this nascent field, by saving fellow researchers significant amounts of time.

\textbf{These did not work!}
\begin{itemize}
    \item Using a single encoder-decoder network to estimate both static and dynamic scene layouts.
    \item Using an ENet~\cite{enet} or a UNet~\cite{unet} architecture as opposed to the ResNet-18 encoder and customized decoder we employed.
    \item Using a DCGAN~\cite{dcgan} as opposed to patch-discriminators \cite{pix2pix}.
    \item Employing a variational autoencoder-style latent code between the encoder and decoder (to allow for sampling)
\end{itemize}

%% file: root.bbl
\begin{thebibliography}{10}\itemsep=-1pt

\bibitem{behley2019iccv}
J.~Behley, M.~Garbade, A.~Milioto, J.~Quenzel, S.~Behnke, C.~Stachniss, and
  J.~Gall.
\newblock Semantickitti: A dataset for semantic scene understanding of lidar
  sequences.
\newblock In {\em ICCV}, 2019.

\bibitem{beltran2018birdnet}
J.~Beltr{\'a}n, C.~Guindel, F.~M. Moreno, D.~Cruzado, F.~Garcia, and
  A.~De~La~Escalera.
\newblock Birdnet: a 3d object detection framework from lidar information.
\newblock In {\em ITSC}, 2018.

\bibitem{ioutracker}
E.~Bochinski, V.~Eiselein, and T.~Sikora.
\newblock High-speed tracking-by-detection without using image information.
\newblock In {\em International Workshop on Traffic and Street Surveillance for
  Safety and Security at IEEE AVSS 2017}, 2017.

\bibitem{psmnet}
J.-R. Chang and Y.-S. Chen.
\newblock Pyramid stereo matching network.
\newblock In {\em CVPR}, 2018.

\bibitem{chang2019argoverse}
M.-F. Chang, J.~Lambert, P.~Sangkloy, J.~Singh, S.~Bak, A.~Hartnett, D.~Wang,
  P.~Carr, S.~Lucey, D.~Ramanan, et~al.
\newblock Argoverse: 3d tracking and forecasting with rich maps.
\newblock In {\em CVPR}, 2019.

\bibitem{chen2016monocular}
X.~Chen, K.~Kundu, Z.~Zhang, H.~Ma, S.~Fidler, and R.~Urtasun.
\newblock Monocular 3d object detection for autonomous driving.
\newblock In {\em CVPR}, 2016.

\bibitem{chen2017multi}
X.~Chen, H.~Ma, J.~Wan, B.~Li, and T.~Xia.
\newblock Multi-view 3d object detection network for autonomous driving.
\newblock In {\em CVPR}, 2017.

\bibitem{cirecsan2012multi}
D.~Cire{\c{s}}an, U.~Meier, and J.~Schmidhuber.
\newblock Multi-column deep neural networks for image classification.
\newblock {\em arXiv preprint}, 2012.

\bibitem{deng2009imagenet}
J.~Deng, W.~Dong, R.~Socher, L.-J. Li, K.~Li, and L.~Fei-Fei.
\newblock Imagenet: A large-scale hierarchical image database.
\newblock In {\em CVPR}, 2009.

\bibitem{KITTI}
A.~Geiger, P.~Lenz, and R.~Urtasun.
\newblock Are we ready for autonomous driving? the kitti vision benchmark
  suite.
\newblock In {\em CVPR}, 2012.

\bibitem{girshick2015fast}
R.~Girshick.
\newblock Fast r-cnn.
\newblock In {\em Proceedings of the IEEE international conference on computer
  vision}, pages 1440--1448, 2015.

\bibitem{godard2018digging}
C.~Godard, O.~Mac~Aodha, M.~Firman, and G.~Brostow.
\newblock Digging into self-supervised monocular depth estimation.
\newblock {\em arXiv preprint}, 2018.

\bibitem{GAN}
I.~Goodfellow, J.~Pouget-Abadie, M.~Mirza, B.~Xu, D.~Warde-Farley, S.~Ozair,
  A.~Courville, and Y.~Bengio.
\newblock Generative adversarial nets.
\newblock In {\em Advances in Neural Information Processing Systems 27}. 2014.

\bibitem{prod}
D.~Hall, F.~Dayoub, J.~Skinner, H.~Zhang, D.~Miller, P.~Corke, G.~Carneiro,
  A.~Angelova, and N.~S{\"u}nderhauf.
\newblock Probabilistic object detection: Definition and evaluation.
\newblock {\em arXiv preprint arXiv:1811.10800}, 2018.

\bibitem{he2017mask}
K.~He, G.~Gkioxari, P.~Doll{\'a}r, and R.~Girshick.
\newblock Mask r-cnn.
\newblock In {\em Proceedings of the IEEE international conference on computer
  vision}, pages 2961--2969, 2017.

\bibitem{he2015delving}
K.~He, X.~Zhang, S.~Ren, and J.~Sun.
\newblock Delving deep into rectifiers: Surpassing human-level performance on
  imagenet classification.
\newblock In {\em ICCV}, 2015.

\bibitem{isola2017image}
P.~Isola, J.-Y. Zhu, T.~Zhou, and A.~A. Efros.
\newblock Image-to-image translation with conditional adversarial networks.
\newblock In {\em CVPR}, 2017.

\bibitem{pix2pix}
P.~Isola, J.-Y. Zhu, T.~Zhou, and A.~A. Efros.
\newblock Image-to-image translation with conditional adversarial networks.
\newblock In {\em CVPR}, 2017.

\bibitem{Adam}
D.~P. Kingma and J.~Ba.
\newblock Adam: A method for stochastic optimization.
\newblock In {\em International Conference on Learning Representatiosn (ICLR)},
  2015.

\bibitem{ku2018joint}
J.~Ku, M.~Mozifian, J.~Lee, A.~Harakeh, and S.~L. Waslander.
\newblock Joint 3d proposal generation and object detection from view
  aggregation.
\newblock In {\em IROS}, 2018.

\bibitem{li2019gs3d}
B.~Li, W.~Ouyang, L.~Sheng, X.~Zeng, and X.~Wang.
\newblock Gs3d: An efficient 3d object detection framework for autonomous
  driving.
\newblock In {\em CVPR}, 2019.

\bibitem{liang2018deep}
M.~Liang, B.~Yang, S.~Wang, and R.~Urtasun.
\newblock Deep continuous fusion for multi-sensor 3d object detection.
\newblock In {\em ECCV}, 2018.

\bibitem{lin2019wise}
S.~Lin, R.~Clark, R.~Birke, N.~Trigoni, and S.~Roberts.
\newblock Wise-ale: Wide sample estimator for aggregate latent embedding.
\newblock 2019.

\bibitem{lu2019monocular}
C.~Lu, M.~J.~G. van~de Molengraft, and G.~Dubbelman.
\newblock Monocular semantic occupancy grid mapping with convolutional
  variational encoder--decoder networks.
\newblock {\em IEEE Robotics and Automation Letters}, 2019.

\bibitem{mousavian20173d}
A.~Mousavian, D.~Anguelov, J.~Flynn, and J.~Kosecka.
\newblock 3d bounding box estimation using deep learning and geometry.
\newblock In {\em CVPR}, 2017.

\bibitem{OpenStreetMap}
{OpenStreetMap contributors}.
\newblock {Planet dump retrieved from https://planet.osm.org }.
\newblock \url{ https://www.openstreetmap.org }, 2017.

\bibitem{enet}
A.~Paszke, A.~Chaurasia, S.~Kim, and E.~Culurciello.
\newblock Enet: A deep neural network architecture for real-time semantic
  segmentation.
\newblock {\em arXiv preprint arXiv:1606.02147}, 2016.

\bibitem{dcgan}
A.~Radford, L.~Metz, and S.~Chintala.
\newblock Unsupervised representation learning with deep convolutional
  generative adversarial networks.
\newblock {\em arXiv preprint arXiv:1511.06434}, 2015.

\bibitem{ren2015faster}
S.~Ren, K.~He, R.~Girshick, and J.~Sun.
\newblock Faster r-cnn: Towards real-time object detection with region proposal
  networks.
\newblock In {\em Advances in neural information processing systems}, pages
  91--99, 2015.

\bibitem{roddick2018orthographic}
T.~Roddick, A.~Kendall, and R.~Cipolla.
\newblock Orthographic feature transform for monocular 3d object detection.
\newblock {\em arXiv preprint}, 2018.

\bibitem{unet}
O.~Ronneberger, P.~Fischer, and T.~Brox.
\newblock U-net: Convolutional networks for biomedical image segmentation.
\newblock In {\em International Conference on Medical image computing and
  computer-assisted intervention}, 2015.

\bibitem{rota2018place}
S.~Rota~Bul{\`o}, L.~Porzi, and P.~Kontschieder.
\newblock In-place activated batchnorm for memory-optimized training of dnns.
\newblock In {\em CVPR}, 2018.

\bibitem{salimans2016improved}
T.~Salimans, I.~Goodfellow, W.~Zaremba, V.~Cheung, A.~Radford, and X.~Chen.
\newblock Improved techniques for training gans.
\newblock In {\em Advances in neural information processing systems}, 2016.

\bibitem{schulter2018learning}
S.~Schulter, M.~Zhai, N.~Jacobs, and M.~Chandraker.
\newblock Learning to look around objects for top-view representations of
  outdoor scenes.
\newblock In {\em ECCV}, 2018.

\bibitem{shi2019pointrcnn}
S.~Shi, X.~Wang, and H.~Li.
\newblock Pointrcnn: 3d object proposal generation and detection from point
  cloud.
\newblock In {\em CVPR}, 2019.

\bibitem{srivastava2019learning}
S.~Srivastava, F.~Jurie, and G.~Sharma.
\newblock Learning 2d to 3d lifting for object detection in 3d for autonomous
  vehicles.
\newblock {\em arXiv preprint}, 2019.

\bibitem{wang2019pseudo}
Y.~Wang, W.-L. Chao, D.~Garg, B.~Hariharan, M.~Campbell, and K.~Q. Weinberger.
\newblock Pseudo-lidar from visual depth estimation: Bridging the gap in 3d
  object detection for autonomous driving.
\newblock In {\em CVPR}, 2019.

\bibitem{wang2019parametric}
Z.~Wang, B.~Liu, S.~Schulter, and M.~Chandraker.
\newblock A parametric top-view representation of complex road scenes.
\newblock In {\em CVPR}, 2019.

\bibitem{yang2018pixor}
B.~Yang, W.~Luo, and R.~Urtasun.
\newblock Pixor: Real-time 3d object detection from point clouds.
\newblock In {\em CVPR}, 2018.

\bibitem{you2019pseudo}
Y.~You, Y.~Wang, W.-L. Chao, D.~Garg, G.~Pleiss, B.~Hariharan, M.~Campbell, and
  K.~Q. Weinberger.
\newblock Pseudo-lidar++: Accurate depth for 3d object detection in autonomous
  driving.
\newblock {\em arXiv preprint}, 2019.

\bibitem{yu2017sketch}
Q.~Yu, Y.~Yang, F.~Liu, Y.-Z. Song, T.~Xiang, and T.~M. Hospedales.
\newblock Sketch-a-net: A deep neural network that beats humans.
\newblock {\em International journal of computer vision}, 2017.

\end{thebibliography}
